 \documentclass[final,5p,times,twocolumn,authoryear]{elsarticle}

\usepackage{amssymb}
\usepackage{lipsum}
\usepackage{booktabs}
\usepackage{multirow}
\usepackage{graphicx}

\makeatletter
\def\ps@pprintTitle{%
  \let\@oddhead\@empty
  \let\@evenhead\@empty
  \let\@oddfoot\@empty
  \let\@evenfoot\@oddfoot
}
\makeatother

\begin{document}

\begin{frontmatter}



\title{
Two-Stage Classifier for Campaign Negativity Detection using Axis Embeddings:\\
\large A Case Study on Tweets of Political Users during 2021 Presidential Election in Iran}


\author[first]{Fatemeh Rajabi}
\affiliation[first]{organization={Amirkabir University of Technology},
            addressline={fateme.rajabi@aut.ac.ir}, 
            city={Tehran},
            country={Iran}}
\author[second]{Ali Mohades}
\affiliation[second]{organization={Amirkabir University of Technology},
            addressline={mohades@aut.ac.ir}, 
            city={Tehran},
            country={Iran}}
\begin{abstract}
In elections around the world, the candidates may turn their campaigns toward negativity due to the prospect of failure and time pressure. In the digital age, social media platforms such as Twitter are rich sources of political discourse. Therefore, despite the large amount of data that is published on Twitter, the automatic system for campaign negativity detection can play an essential role in understanding the strategy of candidates and parties in their campaigns. In this paper, we propose a hybrid model for detecting campaign negativity consisting of a two-stage classifier that combines the strengths of two machine learning models. Here, we have collected Persian tweets from 50 political users, including candidates and government officials. Then we annotated 5,100 of them that were published during the year before the 2021 presidential election in Iran. In the proposed model, first, the required datasets of two classifiers based on the cosine similarity of tweet embeddings with axis embeddings (which are the average of embedding in positive and negative classes of tweets) from the training set (85\%) are made, and then these datasets are considered the training set of the two classifiers in the hybrid model. Finally, our best model (RF-RF) was able to achieve 79\% for the macro F1 score and 82\% for the weighted F1 score. By running the best model on the rest of the tweets of 50 political users that were published one year before the election and with the help of statistical models, we find that the publication of a tweet by a candidate has nothing to do with the negativity of that tweet, and the presence of the names of political persons and political organizations in the tweet is directly related to its negativity.
\end{abstract}




\end{frontmatter}




\section{Introduction}
\label{introduction}

In the information age, where vast amounts of textual data are generated daily, the ability to effectively and efficiently categorize and understand this information has never been more critical than now. Natural Language Processing (NLP) and Machine Learning (ML) have emerged as powerful tools in text classification that promise to automate the understanding of textual information. The task of text classification is not without challenges. Textual data is inherently unstructured and presents issues related to feature extraction, dimensions, and cross-linguistic differences. In addition, with the increase in the volume of data, scalability, efficiency of the model, and other challenges arise. However, within these challenges lie opportunities to provide innovative solutions. ML algorithms, using advanced techniques, promise to discover hidden patterns, sentiments, and insights in textual data \citep{kowsari2019}, \citep{minaee2021}. Recognizing campaign negativity requires a precise understanding of the language and can thus be challenging. Campaign negativity means that the election candidates destroy their competitors instead of presenting their plans, abilities and work histories. This aggressive policy is usually used conceptually and ironically \citep{nai2020}. So far, much work has been done on traditional methods of measuring negativity, such as manual content analysis by political experts, which can be time-consuming and error-prone. To meet this challenge, researchers have turned to automated tools that can quickly and accurately analyze large amounts of data and provide insight into the tone and content of political messages on social media. However, automated tools also have limitations, such as the inability to understand sarcasm and other indirect forms of language. Therefore, it can be important to use a combination of automatic and manual methods to measure the negativity of election campaigns and understand their effects on public opinion and election results.

In previous articles, campaign negativity has been examined from different angles in presidential, senate, municipal, and other elections in different countries. For example, the analysis of the negativity of the campaigns formed in the elections by the candidates, the time when they became negative, the investigation of the causes of the formation of negativity, and the effect of various factors on the campaign negativity have been done \citep{maier2023}, \citep{mattes2020}. The structure of these articles is similar to each other. First, manual analysis or with the help of machine learning algorithms was performed on the desired data set, and then the effect of various variables on the negativity of the campaign was calculated by a statistical regression model. In this article, we tried to present a machine learning model to detect negativity and then answer some hypotheses in the 2021 Iranian presidential election with the help of statistical models.
The challenges we have faced in doing this thesis can be summarized as follows: The first challenge is data collection. Persian tweets published by presidential candidates and a number of political users in Iran were collected through the Twitter Developer Account. After collecting the data, there was a need to manually annotate the tweets, which happened. 5100 tweets from the total collected tweets were labeled into three classes: negativity, personal attacks, and political attacks. Labels were assigned to each class as zero or one. The labeling process is described in Section 3.1.
The next challenge of the problem in the first step is to build a model to predict negativity, for which different classification methods such as classical algorithms and pre-trained deep learning algorithms were used. Considering that detecting the negativity of a tweet is not an easy task even in the tagging stage by humans and sometimes requires reading the tweet several times, in the machine learning model it is also necessary to define the specific features of the problem in order to reach the appropriate accuracy. For this purpose, we have defined features suitable for the type of problem used for the negativity problem (feature extraction). Also, we have used different techniques, like different preprocessing, sequential addition of features, and resampling methods. According to the comparison of different basic ML models, in the proposed solution part, the cosine similarity of embedding of each tweet is used with axis embeddings and a two-stage model, which leads to an increase in the performance of the model. In fact, the challenges we faced during the model can be categorized as follows:
\begin{itemize}
\item Thematic similarity of some positive tweets to negative tweets
\item Using campaign negativity in the form of sarcasm in negative tweets
\item Weakness of embedding models in the Persian language
\end{itemize}

Detecting negativity in an election campaign is a difficult task, and this difficulty was clearly evident in the labeling part of the dataset. In fact, determining the negativity of a tweet that contains one or more short sentences is a difficult task that requires time and money. Therefore, in ML models, it is necessary to know a lot about the type of content and its topics in order to achieve the best performance by defining the appropriate features and choosing the appropriate model.
In the innovation section of this paper, by separating the train dataset into new datasets based on the cosine similarity of embedding tweets with axis embeddings, we were able to significantly improve the accuracy of the model compared to the basic models. We used the average embeddings of positive and negative tweets to create axis embeddings. By setting a threshold for cosine similarity of embedding tweets with average embeddings, we made two subsets of the data set.
In the first subset, we have two classes, positive and negative, where the negative class includes negative tweets in the main data set and some positive tweets in the main data set whose difference between their embedding cosine similarity with the axis embedding 1 (the average embedding of negative tweets) and their embedding cosine similarity with the axis embedding 0 (the average embedding of positive tweets) is greater than the threshold value. We assign these positive tweets Label 2. 
Also, the positive class is made up of the remaining positive tweets in the main data set whose difference between their embedding cosine similarity with the embedding of the axis embedding 1 (the average embedding of negative tweets) and their embedding cosine similarity with the embedding of the axis embedding 0 (the average embedding of positive tweets) is less than the threshold value.
The reason for creating these subgroups is that by calculating the average embedding of negative tweets, it can be seen that some positive tweets are more similar to the average embedding of negative tweets, which is due to the use of irony and sarcasm in negative tweets and the thematic similarity of some positive tweets to negative tweets.
The positive class of the second subset consists of the positive tweets that were labeled 2 in the first subset, whose real label is also positive, and the negative class of this subset is made of the negative tweets of the main dataset, whose real label is also negative.
This method first obtains a general view of how tweets are distributed and how similar they are to each other with the help of axis embeddings to form subsets of the dataset. Then, it uses each subset for the two-stage classification model to determine whether a tweet is negative or not.

In this paper, we first provide an overview of the background and related work in Section 2, which lays the foundation for our research. In Section 3, we present the methodology and data collection process. The results of our study are discussed in detail in Section 4, followed by a conclusion of these results and highlighting the significance of our findings and potential future research directions in Section 5. The subsequent sections of this paper delve into each of these aspects in greater detail. We encourage the reader to follow along to gain a comprehensive understanding of the research presented herein.

\section{Related Words}

\subsection{\textbf{Statistical and Analysis Methods}}
Some methods include an analysis by a political expert to review the content published about the election. Usually, at the end of articles in this category, a statistical model is used to calculate the effect of various factors on the campaign negativity. So far, many articles have used this method to examine various elections in terms of negativity. Here, we have tried to bring articles from 2005 to now that have used different aspects.

In 2005, Peterson et al. investigated the effects of time and party of candidates on their campaign negativity using newspapers related to the 1998 Senate primary elections and concluded that campaign negativity has a dependence on time, party, and number of participants \citep{peterson2005}. 
Krebs et al., in 2007, using newspaper and television ads of candidates in the 2001 Los Angeles mayoral election, aimed to determine whether attacks occurred more on issues or on individuals. They also examine the extent of attacks on minority and non-minority candidates. In the end, they conclude that the issues affect more than the candidate's position on negativity, and minority candidates attack less than non-minorities \citep{krebs2007}. 
In 2009, Schweitzer worked with the aim of comparing the patterns of negativity in Germany and the U.S. by examining the effects of factors such as the number and subject of attacks and whether the candidate is a government candidate or not. He used the candidates' websites in the German national elections and European parliamentary elections for the data set. He finally concluded that the patterns of negativity in German and American electronic campaigns are similar, except for the topics \citep{schweitzer2009}. 
In 2012, Grossmann examined the ads for the 2002 and 2004 US Congressional elections. He wants to examine the causes of negativity in terms of issues and parties. Finally, it is concluded that the incumbent candidates use negative advertisements less than their competitors \citep{grossmann2012}.

In 2016, Hassel et al. conducted a survey of campaign emails in the 2014 U.S. congressional elections to answer the question of when candidates choose to be negative. They conclude that the negativity of emails did not make elections more competitive \citep{hassell2016}.
In a Persian paper in 2018, Babaei et al., by interviewing 16 Iranian political experts, investigated the theoretical foundations of the negative election campaigns from 2004 to 2016 in the Iranian presidential elections. The authors attribute the dominance of the emotional atmosphere, the root of the culture of destruction, the priority of group interests over national interests, and the weakness of laws as the reasons for the negativity of competition \citep{babaei2018}. 
In 2019, Walter et al. sought to increase the validity of measures to identify negative tone in campaigns by analyzing newspapers, voter opinion, and expert opinion in the 2015 UK election and providing a bias removal model. This is a Bayesian statistical model to detect the effect of bias on the opinions of voters and experts. By giving values to the bias variable, its effect is removed in the calculation of negativity \citep{walter2019}. 
In 2020, Maier et al. examined the relationship between negativity and media coverage with statistical models. Analyzing the tweets and TV ads of 507 candidates in 107 national elections in 89 countries from 2016 to 2019. They find that messages with emotional appeal (fear or passion) have a greater impact on media coverage than campaign negativity \citep{maier2020}. 
Nie in 2021 compared the campaign behavior of populists and non-populists by collecting articles on 195 candidates in 40 global elections with election-related keywords from 2016 to 2017. He finds that populists have 15\% more negativity, 11\% more personality attacks, and 8\% more fearmongering messages \citep{nai2021}.

\subsection{\textbf{Machine Learning Methods}}

In 2022, Petkevic et al. presented a model (multilayer perceptron) for detecting negativity and the types of attacks (political or personal) and incivility. The model is trained on 1186 tweets published in the 90 days before the 2018 Senate election by 66 candidates. F1 scores for four classes of negativity, political attacks, personal attacks, and incivility in their best model are 82\%, 83\%, 82\%, and 85\%, respectively. Finally, they used this model to measure the amount of negativity in all 4 classes for 16,000 tweets to measure the effect of various variables such as gender, being Republican, the number of weeks before the election, being in touch with the government, the state of the candidate, and being associated with Trump \citep{petkevic2022} on negativity of campaigns.

In 2023, Kim used tweets related to the 2020 US presidential election and presented a BERT classification model to detect violent tweets. First, he collected stream tweets and then filtered them with political keywords. Then he used a set of violent keywords and filtered the tweets again. Kim claims that many of the tweets contain violent words but do not include attacks. Therefore, by using human labeling for 2500 tweets, he trained different models, the best of which was the BERT model with precision 71.8\%, recall 65.6\%, and F1 score 68.4\%. Then he used this model for active learning technique to tag another 5000 tweets. Finally, he dealt with the negativity binomial regression model to investigate the effect of different factors on violence, which concluded that women and Republicans are more targets of violent tweets than men and non-Republicans \citep{kim2023}.

\section{Methodology}

\subsection{\textbf{Data Collection and Annotation}}
First, the tweets related to 50 political users are collected by get\_all\_tweets endpoint and an appropriate query based on 50 usernames and with a Twitter Developer Account. A total of 42,837 tweets from 50 political users are collected which are published between January 7, 2011, and June 2, 2021. 10,292 tweets are published in the period of one year before the election day, of which 1,776 are by seven candidates. Among these tweets, we have randomly selected and labeled 5100 of them in proportion to the total number of tweets published by each representative. For tagging, the first 3100 tweets were tagged, and then, by running basic models and comparing their accuracy, the best model (ParsBERT) was used for the active learning technique. A number of tweets were given to the ParsBERT model, and about 2000 tweets whose prediction probability was close to the threshold (0.5) (that is, it was difficult for the model to classify them) were selected for tagging. At first, about 500 tweets were tagged by three experts, and then the expert whose tags were most similar to the tags of the majority of the triple tags labeled the rest of the tweets. 

\begin{table}[htb]
\centering
\caption{Number of each label in Dataset (1 shows the tweet has campaign negativity and 0 contrariwise}
\begin{tabular}{|c|c|c|}
\hline
Class/Label & Presence (1) & Absence (0) \\
\hline
Negativity & 1,447 & 3,653 \\
\hline
Political Attack & 507 & 4,593 \\
\hline
Personal Attack & 894 & 4,206 \\
\hline
\end{tabular}
\end{table}
For how to tag a tweet, if the text of the tweet contains a direct or ironic attack or destruction, it will be labeled as 1 for negativity, and otherwise it will be labeled as 0. Also, if the tweet is negative, the type of attack that has been carried out on a person or organization is also determined. Of course, due to the lack of negative tweets, we have not provided a model to detect the type of attack in this article. But in the future, these models can also be formed by adding a negative class of tweets. Table 1 shows the number of tweets in each class.
\subsection{\textbf{Preprocessing and Feature Extraction}}
In supervised classification problems, fixed preprocessing for different problems does not necessarily give the best result. In this regard, we have considered 3 different preprocessing methods for testing models. Finally, we have reported results on the best preprocessing.
\begin{enumerate}
\item \textbf{Preprocessing (1):} converting emojis to text, removing emojis, converting hashtags into separate words, removing repeated characters in a word, removing words including numbers, removing junk characters, removing punctuation, and removing tweets with less than 3 characters (in the Persian language, these tokens don't have meaning).
\item \textbf{Preprocessing (2):} Preprocessing (1) + removing stop words
\item \textbf{Preprocessing (3):} Preprocessing (2) + removing links + removing mentions
\end{enumerate}

In the following, we examine the four categories of features that are defined according to the nature of the data and the type of problem. To test the models, we sequentially add the following sets of features to the classical models and compare the results obtained to see which set of features gives a better answer to the problem. This work can help to understand which categories of features have a better effect on detecting negativity \citep{alizadeh2020}.
\begin{enumerate}
\item \textbf{Text Features: }
The number of retweets, likes, mentions, links, hashtags, insulting words, names of organizations and political persons, sentiment analysis, ...
\item \textbf{Metatext Features: }
Embedding tweets, unigram, bigram and frequent trigrams, frequent class 1 and 0 tokens with and without subscription, ...
\item \textbf{User Features: }
Number of followers, followings, likes, tweets, most frequent tokens in descriptions, ...
\item \textbf{Time Features: }
The lifespan of the user's page, the publication of tweets in 4 intervals of the day and night, the difference in the time of publication of tweets in the intervals of 10, 20, ... days before the election, ...
\end{enumerate}
In total, we have defined about 1400 features for classical models.

\subsection{\textbf{Building New Datasets}}
Now we want to present the method for building the necessary data sets for the two-stage model. For this purpose, we use two methods: axis embeddings and clustering. Before explaining the mentioned methods, it should be mentioned that according to the results of the basic models (classical and deep learning) that we have presented in the results section, for building new datasets we used the ALC embedding, which was the most accurate. This method replaces the usual Word2Vec method and is based on GloVe embeddings and a linear transformation. It is also suitable for rare words in the corpus. In the related paper on ALC embedding, the authors claim that the ALC model needs fewer examples for learning than other models. Also, the quality of ALC embeddings has been better than other models in many examples.\citep{khodak2018}

\subsubsection{\textbf{Axis Embeddings Trick}}
Before dealing with the creation of new labels and subsets of the dataset, it is necessary to understand the role of axis embeddings in text classification. Axis embeddings in this paper are the representation of documents in a continuous vector space. These axis embeddings  are calculated by averaging the embeddings of tweets in different classes. We define two axis embeddings:
\begin{itemize}
    \item \textbf{Axis-embedding 1 (EMB1):}  represents the average embedding of tweets labeled as class 1 (negative), which means the tweet contains campaign negativity.
    \item \textbf{Axis-embedding 0 (EMB0):} represents the average embedding of tweets labeled as class 0 (positive), which means that the tweet has no campaign negativity.
\end{itemize}

By converting the semantic content of tweets into continuous vectors, we are able to quantify the similarity between tweets and axis representations. In fact, based on this similarity, we form new tags and subcategories.

The innovation presented in this section revolves around identifying tweets in class 0 that are more similar to EMB1. These tweets are often about the topics of negative tweets that have been sarcastically discussed. In our proposed model, we classify them as negative, since text features are more important than other features in this problem (according to the feature importance of basic models). In other words, label 2 is introduced for these tweets.  Tweets with label 2 have a difference in similarity with EMB1 and their similarity with EMB0 is more than the threshold.\citep{zhou2022} For each tweet with label 2, the formula (2) is true. In this condition, \textit{CS} refers to the cosine similarity based on formula (1) between 2 vectors, \textit{X} refers to the ALC embedding of tweets in the train set, and \textit{t} refers to the threshold.
\begin{equation}
    CS(A, B)=\frac{A.B}{|A|.|B|}
\end{equation}
\begin{equation}
CS(X, EMB1)-CS(X, EMB0)> t
\end{equation}
\begin{enumerate}[1.]

\item \textbf{New Dataset 1:} The first new data set (train set for first classification in the two-stage model) consists of tweets labeled zero and one, which are considered as follows.
\begin{enumerate}
\item Positive tweets (labeled positive in the original data set) that do not include negativity, and the difference of the cosine similarity of their embedding with EMB1 and the cosine similarity of their embedding with EMB0 is less than the threshold. The negativity label for this class is considered zero.
\item Positive tweets (labeled positive in the original data set) that do not include negativity and the difference between the cosine similarity of their embeddings with EMB1 and their cosine similarity of their embeddings with EMB0 is greater than the epsilon (wherever label 2 is mentioned, it means this group of tweets). The negative label of this class is considered one. In fact, the embedding of these tweets is closer to the average embedding of negative tweets. (due to the similarity in topics with negative tweets and using sarcasm in negative tweets)
\item Negative tweets (labeled one in the original dataset) that contain negativity. (It should be noted that all these tweets are closer to EMB1)
\end{enumerate}
\item \textbf{New Dataset 2:} The second new dataset (train set for second classification in the two-stage classifier) consists of tweets labeled zero and one, which are considered as follows.
\begin{enumerate}
\item Positive tweets (labeled zero in the original data set) that do not contain negativity and the difference between the cosine similarity of their embeddings with EMB1 and their cosine similarity of their embeddings with EMB0 is greater than the epsilon (we defined it with label 2 in the previous section). The negativity label for this class is considered zero.
\item Negative tweets (labeled negative in the original dataset) that contain negativity.
\end{enumerate}
\end{enumerate}

\subsubsection{\textbf{Clustering Trick}}
In this section, instead of using axis embeddings to construct new datasets for the two-stage model, we use different clustering methods to separate tweets. Here we use DBSCAN \citep{ouyang2022online}, K-means \citep{ikotun2022k}, Agglomerative \citep{randriamihamison2021applicability}, Birch \citep{ramadhani2020improve}, Gaussian Mixture (GM) \citep{adams2019survey}, and Optics \citep{bhattacharjee2021survey} clustering methods.
Each of these clustering methods has its strengths and weaknesses, and the choice of which one to use depends on the specific characteristics of the data and the objectives of the clustering task. It is often a good idea to try several methods and compare their results to find the most suitable clustering approach for a particular data set.
Here, the difference in building the data set based on clustering methods instead of the axis embedding trick is in separating tweets with tag 2. In this trick, the tweets that are placed in the same cluster as the majority of negative tweets are considered as class 2, and the rest of the positive tweets are considered as class 0, and similarly to the previous section, the new datasets will be made.

\subsection{\textbf{Two-Stage Model}}
Creating these new datasets serves a dual purpose. First, it addresses the challenge of classifying tweets that exhibit characteristics that vary by topic. Second, due to the combination of some positive tweets with negative tweets in the first classifier as class one, they will be predicted by the second classifier to be identified this time based on their real label. This innovative approach increases the power of the model in detecting the negativity of election campaigns and contributes to a deeper understanding of the complexities of analyzing negativity in the political landscape.
In fact, we have a hybrid classification model that works in two separate stages. In the first stage, the first classifier model is trained with the entire available training data (the first new dataset) to develop a comprehensive understanding of the patterns in the text that are relevant to election campaigns. At this early stage, the foundation of the classifier's ability to effectively identify and classify tweets that do not contain campaign negativity and are not conceptually similar to negative tweets is laid. In the second stage, the second classification model is trained on the second new dataset, and re-prediction is performed focusing on the tweets predicted as class one (negative) in the first model. Apparently, following this process will determine the real tag of positive tweets, which are similar to negative tweets. This two-stage approach not only contributes to a more accurate initial classification but also to increased overall accuracy in detecting campaign negativity. In Figure 1, you can see the final structure of the model. This figure generally includes the dataset preparation (left section) and the two-stage model architecture (right section).
\begin{figure}[htb]
\centering
\includegraphics[width=0.5\textwidth]{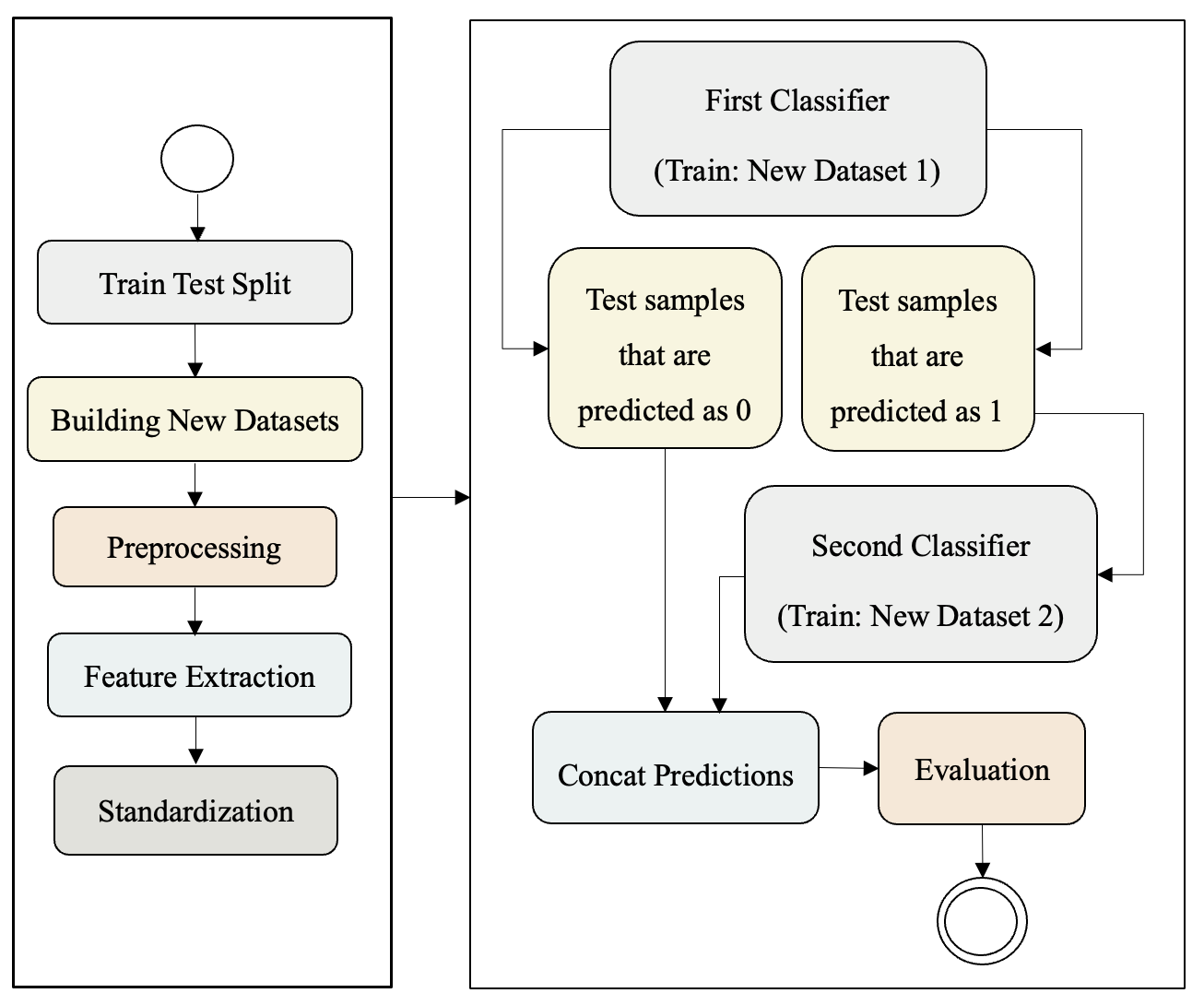}
\caption{Architecture of Two-Stage Classifier (Proposed Model(left section shows dataset preparation and right section shows the two-stage model architecture}
\end{figure}

\section{Results}
\subsection{\textbf{Evaluation Metrics}}
The process of evaluating machine learning models is essential to understanding their performance, reliability, and suitability for real-world applications. This chapter discusses the various evaluation models, techniques, and metrics that are essential for comprehensively evaluating the performance of machine learning algorithms. Model evaluation not only helps researchers and practitioners make informed decisions but also plays an important role in advancing new machine-learning methods \citep{tharwat2020classification}, \citep{li2022survey}. In the results section, we examine the F1-score, precision (P), and recall (R) for two classes (positive and negative), F1-macro and F1-weighted, which are listed in formulas (3), (4), (5), (6), and (7), respectively. Because the F1-score is a combination of two criteria: precision and recall, both of which are important in this issue. In formulas (3), (4), and (5) \textit{TN} refers to the number of true negative labels,\textit{TP} refers to the number of true positive labels, \textit{FN} refers to the number of false negative labels, and \textit{FP} refers to the number of false positive labels (positive and negative are based on case study class). In formulas (4) and (5), \textit{N} refers to the number of classes (here it is 2).
\begin{equation}
    Precision (P) = \frac{TP}{TP+FP}
\end{equation}
\begin{equation}
    Recall (R) = \frac{TP}{TP+FN}
\end{equation}
\begin{equation}
    F1 = \frac{2TP}{2TP+FN+FP} = \frac{2\times Precison\times Recall}{Precision + Recall}
\end{equation}
\begin{equation}
    {F1}_{macro} = \frac{1}{N} \sum_{i=1}^{N} {F1}_i
\end{equation}
\begin{equation}
    {F1}_{weighted} = \sum_{i=1}^{N} {w}_i \times {F1}_i
\end{equation}
\subsection{\textbf{Basic models}}
In this section, we first examine the results related to the basic models and then compare the best results with the results related to the two-stage model with axis embedding and clustering tricks. In the basic models, after separating the train and test sets in a ratio of 85 to 15, stratifying on the negativity label and data preparation, we test the models on different preprocessings and by sequentially adding the features that we discussed in Section 3.2. Also, due to the inequality of the number of samples in the two positive and negative classes in the dataset, we use Smote and TomekLink techniques. The Smote method is an oversampling method that generates artificial samples for the minority class. It does this by creating synthetic instances that are combinations of existing minority-class instances. It addresses the overfitting problem associated with random oversampling by generating new and diverse data points. The tomeklink method involves separating the samples into pairs (one from the majority class and one from the minority class) that are close to each other but from different classes. Removing the majority of class samples in these pairs can help improve the separation between classes. This method can be used for downsampling to lead to better separation of classes without introducing artificial data.
Basic models include classic models and pretrained Deep Learning models (suitable for the Persian language). Classical models are multilayer perceptron (MLP) \citep{liu2022we}, eXtreme Gradient Boosting (XGB) \citep{gohiya2018survey}, Random Forest (RF) \citep{resende2018survey}, Logistic Regression (LR) \citep{boateng2019review}, Naive Bayes (NB) \citep{wickramasinghe2021naive}, Support Vector Machine (SVM) \citep{chandra2021survey}, Gaussian Naive Bayes (GNB) \citep{wijayanto2018experimental}, K-Nearest Neighbor (KNN) \citep{sha2020knn}, Ridge \citep{dobriban2018high}, Gradient Boosting (GB) \citep{sun2020gradient}, and Stochastic Gradient Descent (SGD) \citep{mignacco2020dynamical}. Pretrained Deep Learning models include ParsBERT (DistilBERT-ZWN), ParsBERT (BERT-ZWN) \citep{farahani2021parsbert}, Multilingual DistilBERT \citep{sanh2019distilbert} and Multilingual BERT (MBERT) \citep{pires2019multilingual} which are trained on Persian. For classic models, we have used various embeddings such as FastText, Word2Vec \citep{thavareesan2020sentiment}, GloVe \citep{dharma2022accuracy} and ALC. These embeddings are considered metatext features. According to Table 2, the results of the classical RF model with ALC embedding indices have performed better than the rest of the classic and deep models (the best model of the deep models is MBERT) which can be considered the reason for the strong performance of the ALC model in embedding words with low repetition. Because there are many rare words in the texts in our dataset. Also, the results of model RF (ALC) have been obtained on preprocessing (3) and with all features (text features, metatext features, user features, and time features). Additionally, model MBERT has the best results with preprocessing (3), which are reported in Table 2.

\begin{table}[htb]
\centering
\caption{Evaluation metrics values for best classic and deep models (class 1 is negative tweets and class 0 is positive tweets)}
\label{table2}
\resizebox{.46\textwidth}{!}{%
\begin{tabular}{@{}ccccccc@{}}
\toprule
Model                    & Class & P    & R    & F1   & F1m                   & F1w                   \\ \midrule
\multirow{2}{*}{RF(ALC)} & 1     & 72\% & 50\% & 60\% & \multirow{2}{*}{72\%} & \multirow{2}{*}{78\%} \\
                         & 0     & 81\% & 92\% & 86\% &                       &                       \\ \midrule
\multirow{2}{*}{MBERT}   & 1     & 54\% & 70\% & 61\% & \multirow{2}{*}{70\%} & \multirow{2}{*}{74\%} \\
                         & 0     & 85\% & 75\% & 80\% &                       &                       \\ \bottomrule
\end{tabular}%
}
\end{table}

Figure 2 shows how tweets are distributed using their ALC embedding, which addresses our paper challenge related to the semantic proximity of some positive tweets to negative tweets. In this figure, by using two-dimensionality reduction methods, PCA and TSNE \citep{pareek2021data}, it can be seen that some positive tweets are very close to negative tweets. It seems that some positive tweets have fallen on top of negative tweets. Of course, this is only a representation of the reduced dimensions of tweet embeddings. Therefore, in the next section, we want to implement a two-stage method using axis embedding and clustering tricks.
\begin{figure}[htb]
\centering
\includegraphics[width=0.49\textwidth]{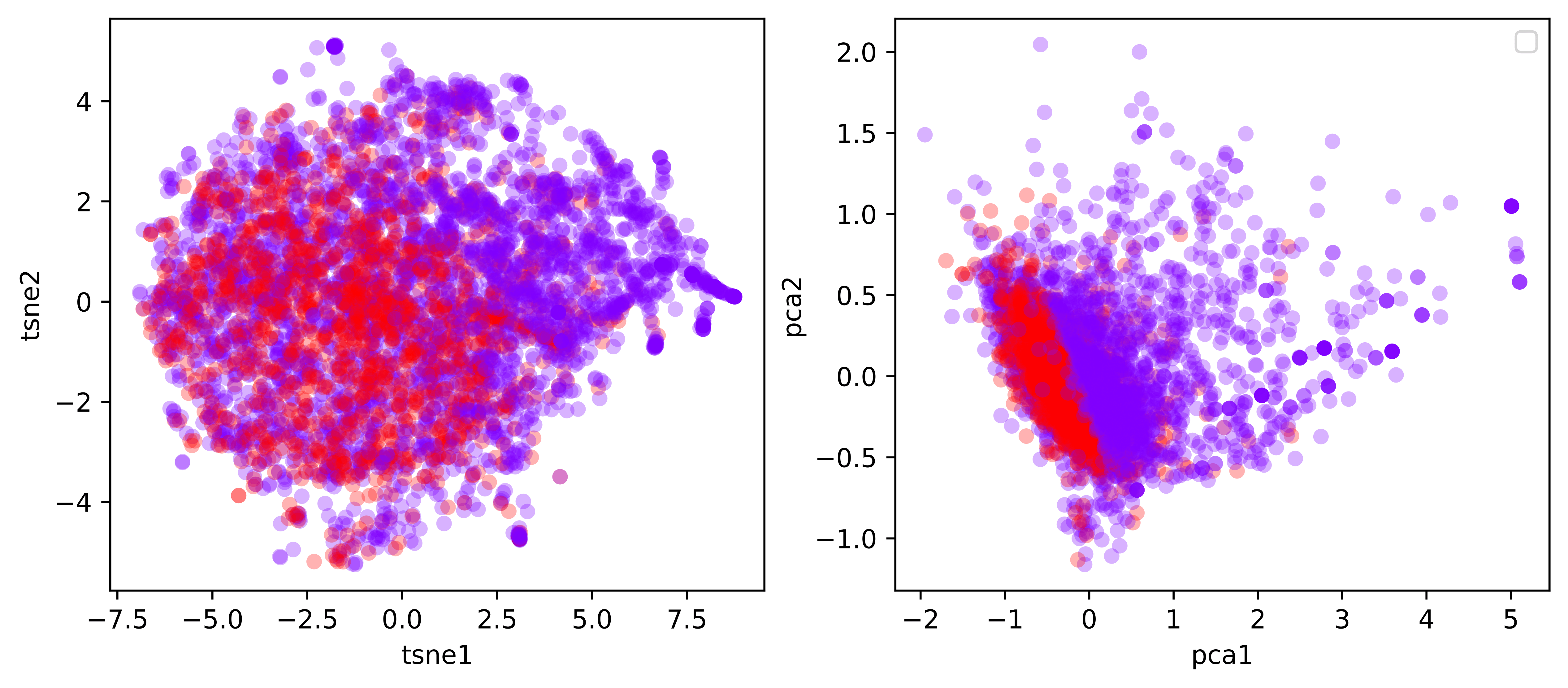}
\caption{Labels of tweets are shown by TSNE and PCA methods for ALC embedding of tweets (purple color shows positive tweets and red color shows negative tweets)}
\end{figure}

\subsection{\textbf{Our approach}}
In this section, we use two clustering methods (Gaussian Mixture and Birch) and three threshold values (0, 0.03, and 0.05) for the axis embedding trick, to create new data sets. The reason for using the two mentioned clustering methods is that they did not separate the tweets that included negativity as much as possible, and their focus was to cluster positive tweets in line with our goal. Based on Figure 3, which shows the clustering of each of the methods according to their ALC embeddings. DBSCAN and Optics methods did not perform well. KMeans and Agglomerative methods have done more separation in class 1 (compared to Figure 2), which is not our goal. But the Gaussian Mixture and Birch methods have focused their separation on positive tweets and put negative tweets in a cluster as much as possible.

\begin{figure}[htb]
\centering
\includegraphics[width=0.49\textwidth]{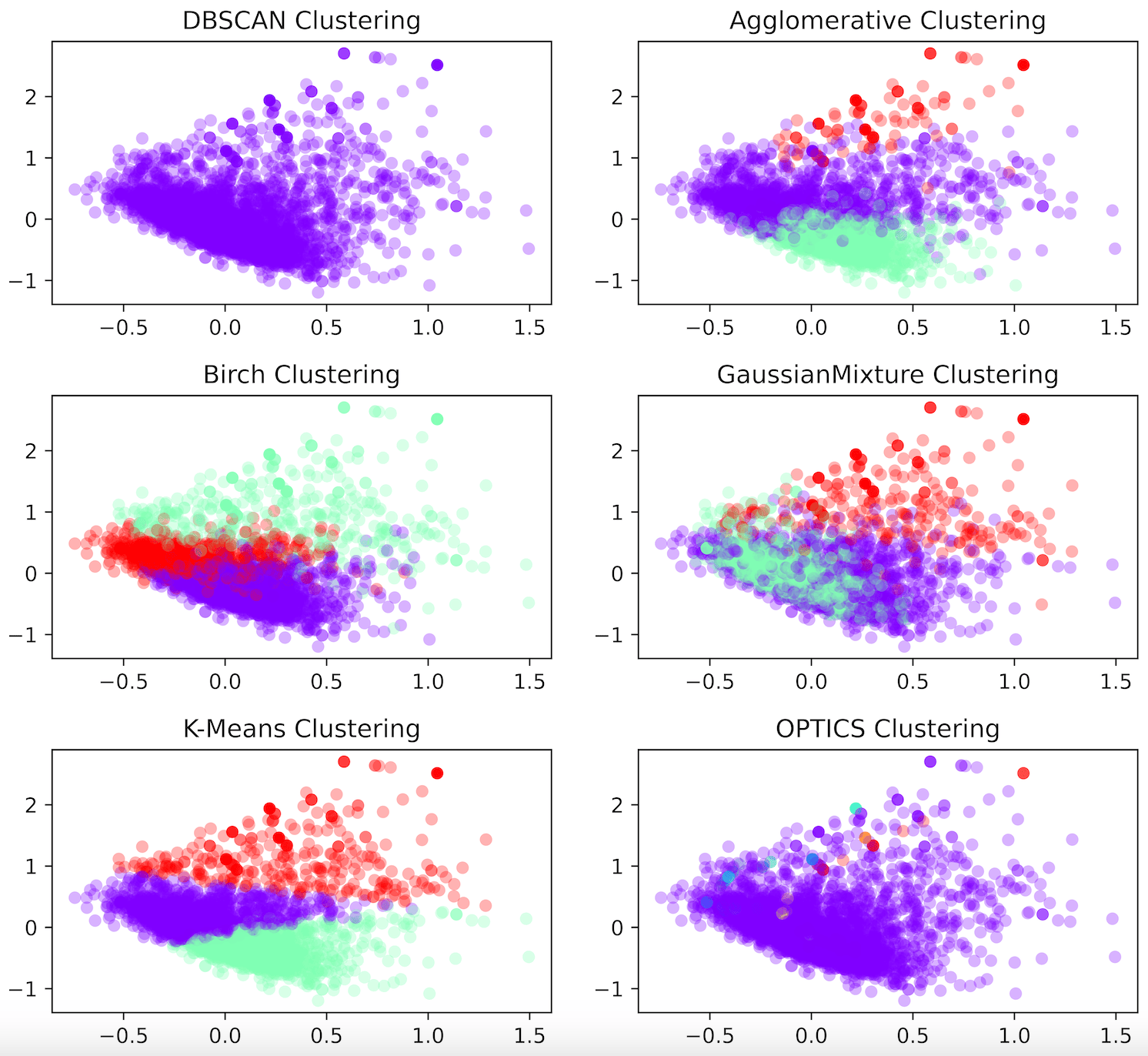}
\caption{The embedding distribution of tweets in dimensionality reduced by PCA with different clustering methods.)}
\end{figure}

Table 3 shows the number of labels in the new datasets created with the five mentioned tricks. In this table, 4335 tweets are included in the train collection and 765 tweets are included in the test collection (85 to 15 ratio). Also, in the axis embedding methods, the number of Label 2 has been reduced at higher threshold values. Because according to formula (2), with an increase in threshold value, less weight is given to positive tweets to be similar to the average of negative tweets. As can be seen in Table 3, in all methods, the number of negative tweets is fixed equal to 1230, which is the total number of negative tweets in the training set.

\begin{table}[htb]
\centering
\caption{Number of each label in new datasets with 5 methods(axis embedding and clustering)}
\label{table1}
\resizebox{.44\textwidth}{!}{
\begin{tabular}{lcccc}
\hline
\multicolumn{1}{c}{\multirow{2}{*}{Method}} & \multicolumn{4}{c}{Train Set} \\ \cline{2-5}
\multicolumn{1}{c}{} & label 0 & label 1 & label 2 & sum \\ \hline
\multicolumn{1}{l|}{Axis\_Emb (t=0)} & \multicolumn{1}{c|}{1219} & \multicolumn{1}{c|}{1230} & \multicolumn{1 }{c|}{1886} & 4335 \\
\multicolumn{1}{l|}{Axis\_Emb (t=0.03)} & \multicolumn{1}{c|}{1695} & \multicolumn{1}{c|}{1230} & \multicolumn{1 }{c|}{1410} & 4335 \\
\multicolumn{1}{l|}{Axis\_Emb (t=0.05)} & \multicolumn{1}{c|}{2213} & \multicolumn{1}{c|}{1230} & \multicolumn{1 }{c|}{892} & 4335 \\
\multicolumn{1}{l|}{Clustering (GM)} & \multicolumn{1}{c|}{1260} & \multicolumn{1}{c|}{1230} & \multicolumn{1}{c| }{1845} & 4335 \\
\multicolumn{1}{l|}{Clustering (Birch)} & \multicolumn{1}{c|}{1155} & \multicolumn{1}{c|}{1230} & \multicolumn{1}{c| }{1950} & 4335 \\ \hline
\end{tabular}}
\end{table}

Now, in Table 4, we see the results of the best hybrid models for the proposed two-stage classification method. In this table, all the basic models considered in Section 4.2 in the first and second stages of our approach have been tested, and the best models have been selected in terms of macro F1 score. Finally, with the consensus of the predictions of the first and second classifications, the best model is RF-RF with the axis embedding trick with threshold 0, whose F1-macro is 79\% and F1-weighted is 82\%. Also, the Birch method has been able to be very close to the best accuracy.

\begin{table}[htb]
\caption{Best results of two-stage models in different methods for building new datasets}
\label{tab:my-table}
\resizebox{.49\textwidth}{!}{%
\begin{tabular}{@{}lccccccc@{}}
\toprule
\multicolumn{1}{c}{Method} & Model & Class & P      & R      & F1     & F1m & F1w \\ \midrule
\multirow{2}{*}{t=0}    & \multirow{2}{*}{RF-RF}   & 1 & 70.3\% & 66\%   & 75.2\% & \multirow{2}{*}{78.8\%} & \multirow{2}{*}{81.9\%} \\
                           &       & 0     & 87.3\% & 89.8\% & 84.9\% &     &     \\ \midrule
\multirow{2}{*}{t=0.03} & \multirow{2}{*}{MLP-GNB} & 1 & 57.1\% & 42.9\% & 85.5\% & \multirow{2}{*}{60.1\%} & \multirow{2}{*}{59.4\%} \\
                           &       & 0     & 63\%   & 88.2\% & 48.9\% &     &     \\ \midrule
\multirow{2}{*}{t=0.05} & \multirow{2}{*}{SVM-LR}  & 1 & 72.3\% & 82.6\% & 77.1\% & \multirow{2}{*}{32.9\%} & \multirow{2}{*}{65\%}   \\
                           &       & 0     & 52.6\% & 51.9\% & 51.4\% &     &     \\ \midrule
\multirow{2}{*}{GM}     & \multirow{2}{*}{MLP-GNB} & 1 & 51.8\% & 50.7\% & 52.9\% & \multirow{2}{*}{66.5\%} & \multirow{2}{*}{72.7\%} \\
                           &       & 0     & 81.1\% & 81.8\% & 80.5\% &     &     \\ \midrule
\multirow{2}{*}{Birch}  & \multirow{2}{*}{LR-XGB}  & 1 & 68.4\% & 72.8\% & 64.5\% & \multirow{2}{*}{77.2\%} & \multirow{2}{*}{80.9\%} \\
                           &       & 0     & 85.9\% & 83.7\% & 88.3\% &     &     \\ \bottomrule
\end{tabular}%
}
\end{table}
\section{Discussion}
As we said in Section 3.1, 10,292 tweets were published by 50 users within a year before election day. Now we run the best RF-RF model on 5,192 unlabeled tweets to identify their negativity label. In total, we have 3,258 negative tweets and 7,034 positive tweets. Now, using a negative binomial regression model, we examine the effect of various factors on negativity. In this model, we have an independent variable is\_candidate that shows whether a user is a candidate or not in the election. As can be seen in Table 5, this variable is not at a statistically significant level, which means that there is no relationship between being a candidate and negative tweets in the 2021 Iranian presidential election. Also, the person\_names\_count variable, which shows the number of political persons' names used in tweets, is at a statistically significant level and its coefficient value is high, which means that there is a direct relationship between the presence of political persons' names in tweets and its negativity.
\begin{table}[htb]
\caption{Drivers of campaign negativity in 2021 presidential election in Iran \\($*:p \le 0.1, **:p \le 0.05, ***:p \le 0.01$)}
\label{undefined}
\resizebox{.44\textwidth}{!}{%
\begin{tabular}{l|c|c|c}
\hline
variable names             & coef    & std err & p     \\ \hline
tweet\_age                 & -0.0001 & 1.0e-4  & 0.273 \\
account\_age ***           & -0.0002 & 1.0e-4  & 0.003 \\
organize\_names\_count *** & 0.1348  & 4.5e-2  & 0.003 \\
person\_names\_count ***   & 0.2175  & 6.8e-2  & 0.001 \\
is\_candidate              & 0.0197  & 9.7e-2  & 0.839 \\
swear\_words\_count        & 0.1146  & 1.7e-1  & 0.5   \\ \hline
\end{tabular}%
}
\end{table}
It can also be seen that the variables related to the life of the tweet and the number of swear words in the tweets are not at a significant level. The user account life variable is also at a statistically significant level, but it has an indirect relationship with negativity with a low coefficient. The variable count of organization names is at a statistically significant level, and its relationship with negativity is positive. This means that tweets that include the names of government organizations and institutions have a high probability of being negative.
\section{Conclusion}
Our goal in this paper was to present an automatic model to detect campaign negativity, which we discussed in the introduction. One of the most important reasons for creating this model is the faster identification of negativity in the campaigns formed in an election by the parties, which can give better and faster answers to the attacks formed by the opposite parties. This can also help in making more accurate and faster reports by different media and be used as a component in social media analysis dashboards by private companies. One of the most important motivations for this paper is the failure to do similar work for Persian data related to Iran's elections. For the first time, we were able to present an artificial intelligence model in this direction. One of the things that can be done in the future following this article is the use of Large Language Models (LLM). In addition to labeling new data and improving the accuracy of the model, these models can also be used as separate models, and their results can be compared with the results of current models by optimizing prompts and executing them on experimental data. Among other things that can be added to the current model is the ability to recognize the type of attack in terms of political attack and personal attack, which requires increasing data that includes negativity. Another suggestion is to add the ability to recognize text from images because many tweets include an image with text.

\section*{Acknowledgments}

I would like to express my sincere gratitude to my advisor, Meysam Alizadeh, for their guidance, mentorship, and continuous support throughout this research project.

\bibliographystyle{elsarticle-harv} 
\bibliography{main}






\end{document}